\documentclass[10pt,twocolumn,letterpaper]{article}

\usepackage[pagenumbers]{cvpr} 

\usepackage{tabularx}
\usepackage{float}
\usepackage{lipsum}
\usepackage{placeins}

\definecolor{cvprblue}{rgb}{0.21,0.49,0.74}
\usepackage[pagebackref,breaklinks,colorlinks,allcolors=cvprblue]{hyperref}
\usepackage{marvosym}

\title{Realistic Corner Case Generation for Autonomous Vehicles with Multimodal Large Language Model}

\author{Qiujing Lu \quad Meng Ma \quad Ximiao Dai \quad Xuanhan Wang \quad Shuo Feng \thanks{Corresponding author: fshuo@mail.tsinghua.edu.cn}
\\
Tsinghua University}

\begin{document}
\maketitle
\begin{abstract}
To guarantee the safety and reliability of autonomous vehicle (AV) systems, corner cases play a crucial role in exploring the system's behavior under rare and challenging conditions within simulation environments. However, current approaches often fall short in meeting diverse testing needs and struggle to generalize to novel, high-risk scenarios that closely mirror real-world conditions. To tackle this challenge, we present AutoScenario, a multimodal Large Language Model (LLM)-based framework for realistic corner case generation. It converts safety-critical real-world data from multiple sources into textual representations, enabling the generalization of key risk factors while leveraging the extensive world knowledge and advanced reasoning capabilities of LLMs.Furthermore, it integrates tools from the Simulation of Urban Mobility (SUMO) and CARLA simulators to simplify and execute the code generated by LLMs. Our experiments demonstrate that AutoScenario can generate realistic and challenging test scenarios, precisely tailored to specific testing requirements or textual descriptions. Additionally, we validated its ability to produce diverse and novel scenarios derived from multimodal real-world data involving risky situations, harnessing the powerful generalization capabilities of LLMs to effectively simulate a wide range of corner cases.
\end{abstract}
\section{Introduction}
\label{sec:intro}
Currently, the safety of autonomous vehicles (AVs) remains a critical barrier to their widespread deployment on public roads. Discovering and testing corner cases in advance helps secure AV safety and accelerating development cycles. However, as AV performance improves, further advancements become increasingly difficult. This is due to corner cases emerging less frequently and exhibiting greater diversity \cite{liu2024curse}. As a result, defining and identifying the most relevant corner cases has become increasingly critical for achieving further performance gains. 

Significant efforts have been made in this area. For instance, CODA\cite{li2022coda} carefully mined corner cases from large-scale autonomous driving datasets\cite{caesar2020nuscenes,geiger2012we,mao2021one}. However, this approach is limited by its reliance on real-world driving data collected from AVs, which is both costly and constrained in scope. Additionally, replaying pre-collected data lacks dynamic interaction with the AV under test, reducing its effectiveness. On the other hand, various methods have been explored to synthesize safety-critical scenarios. These include rule-based \cite{Rana_Malhi_2021, Althoff_Lutz_2018} and data-driven techniques \cite{wang2021advsim, Ding_Xu_Zhao_2020, rempe2022generating}. However, these methods often suffer from limited diversity due to their dependence on initial conditions from given scenes. Furthermore, the scenarios generated through predefined rules or adversarial learning may lack realism, as the applied perturbations can deviate from plausible real-world behaviors, thereby diminishing the effectiveness of the testing process.

\begin{figure}
  \centering
\includegraphics[width=0.48\textwidth]{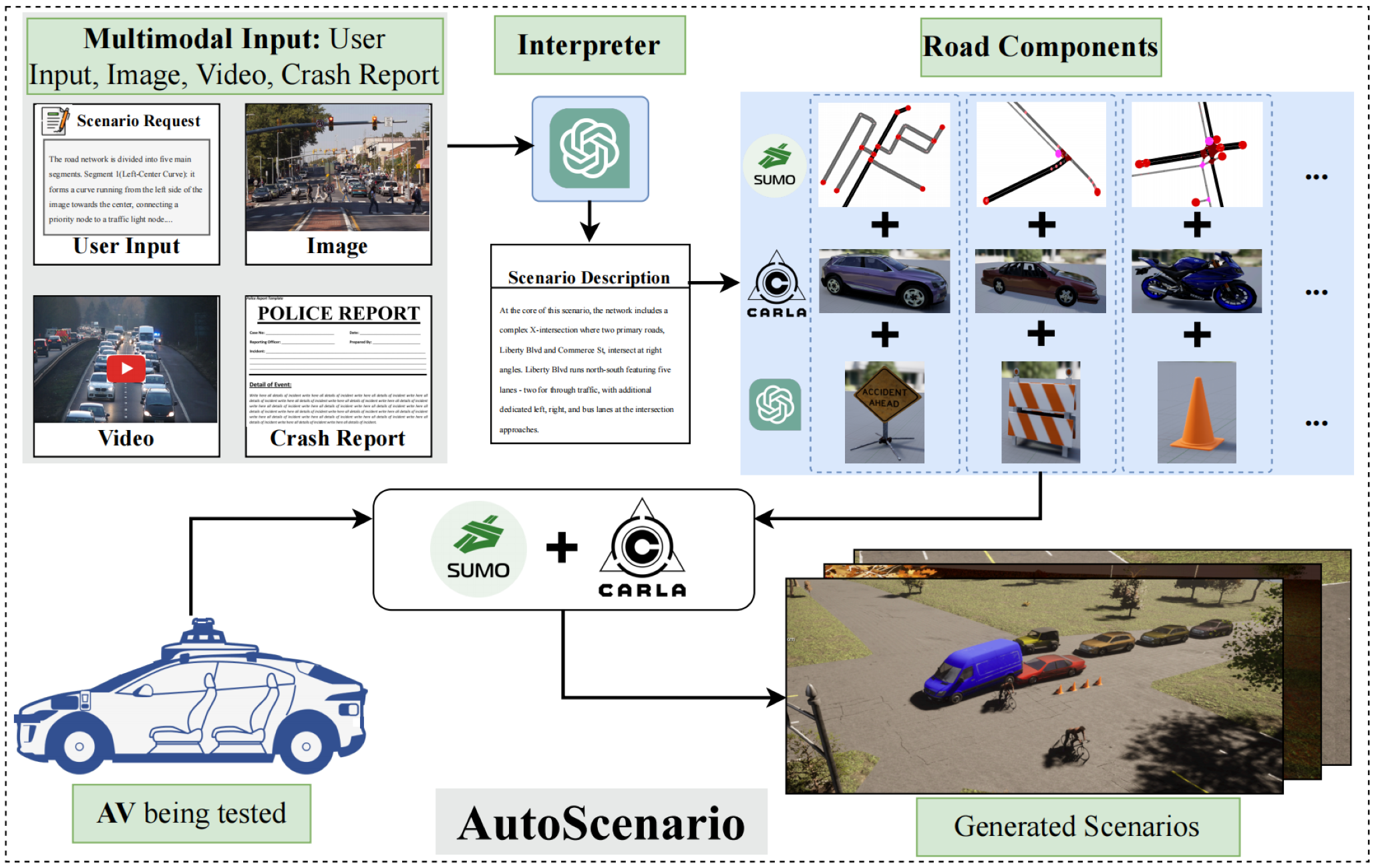}
  \caption{AutoScenario: an LLM based framework for automated generation of realistic corner cases. }
\label{fig:intro}
\end{figure}

Meanwhile, there has been limited progress in developing effective control mechanisms for flexible scenario generation based on abstract requirements. This is crucial as developers often conceptualize scenarios broadly, whereas simulations require detailed configurations, such as road geometry and precise vehicle placements. A mechanism that enables developers to control scenario generation through language descriptions provides a natural solution to bridge this gap, making scenario-based testing more practical and accelerating performance evaluation of AVs.

However, building such text-conditioned generation mechanism is challenging as it requires modeling everything from static environment elements to agent behaviors while mapping narrative language to detailed configurations. The rise of LLMs and Vision-Language Models (VLMs), trained on vast internet data, offers a promising approach, as it has been shown exceptional capabilities in learning, reasoning, and complex problem-solving. Their applications span fields like medicine, education, finance, and engineering \cite{chen2021evaluating,li2022competition,imani2023mathprompter,kumaran2023scenecraft}, showcasing significant advancements.

Driven by these advancements and the need for realistic and diverse safety- critical scenarios, we developed AutoScenario, a fully automated pipeline with high controllability as shown in Fig. \ref{fig:intro}. It generates realistic and diverse scenarios containing main components that closely mimic real-world environments through prompt engineering and the integration of tools from SUMO, an open-source traffic simulation package \cite{lopezMicroscopicTrafficSimulation2018}, CARLA\cite{dosovitskiy2017carla}, an open-source simulator powered by Unreal Engine \cite{unrealengine} with high-fidelity digital assets, and data-driven deep learning models. 

\begin{figure*}[ht!]
  \centering
\includegraphics[width=0.95\textwidth]{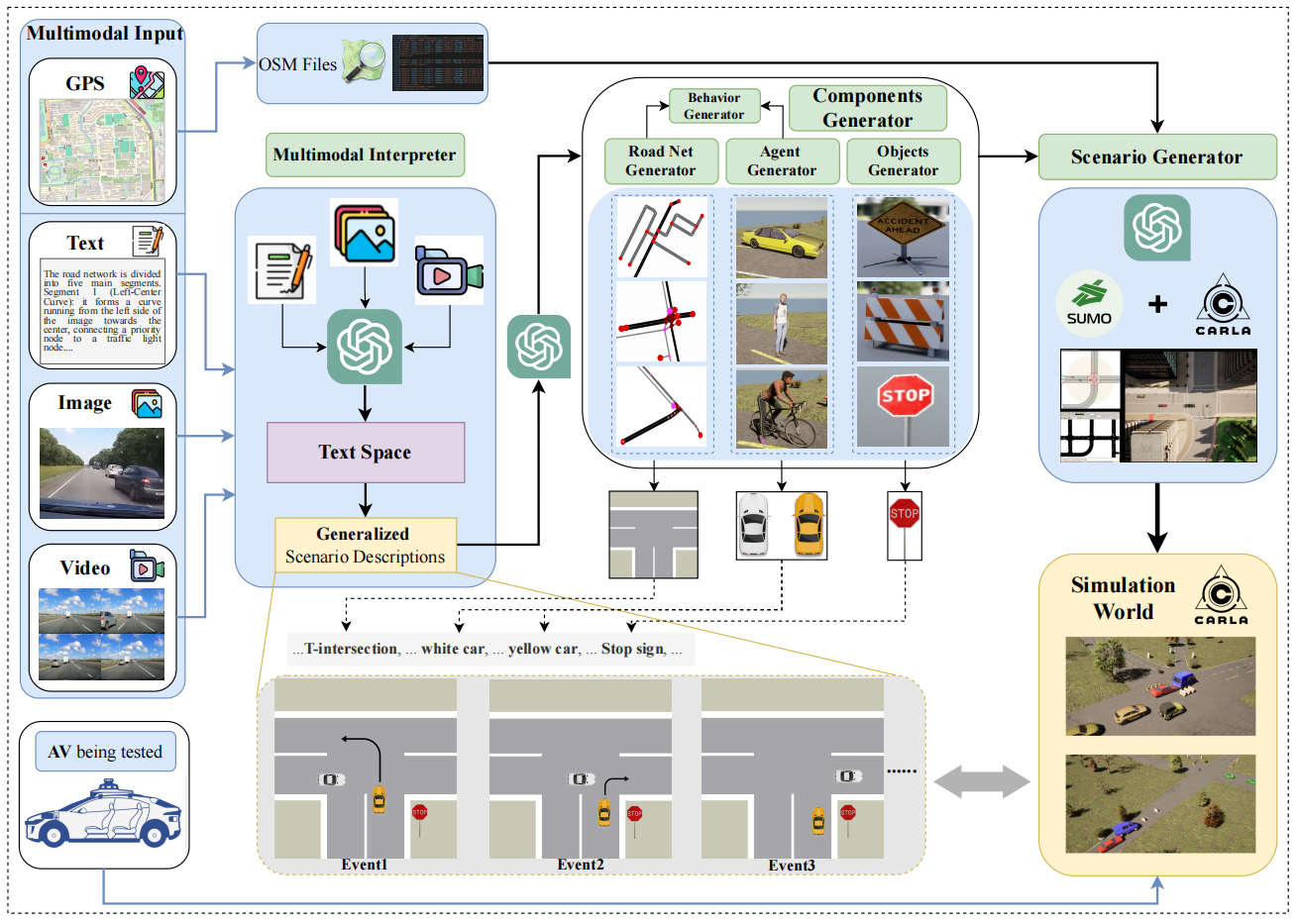}
  \caption{AutoScenario system overview: it accepts multimodal inputs, which are processed by the Multimodal Interpreter. Based on the generalized scenario description, the Components Generator activates to build key components, after which the Scenario Generator is used for scenario testing..}
\label{fig:framework}
\end{figure*}

Our contribution can be concluded as :
\begin{itemize}
\item We propose AutoScenario, a framework that automates the safety-critical scenario generation pipeline while providing a high degree of controllability.
\item Multimodal real-world corner cases are efficiently leveraged to enhance the realism of generated scenarios while preserving key risk factors.
\item We utilize large language model to generalize scenarios through reasoning and open-world knowledge and employ simulation tools to increase the stability and realism of generated scenarios.
\end{itemize}

\section{Related works}
\label{sec:related_work}

\subsection{Safety-Critical Driving Scenario Generation}

The widespread deployment of autonomous vehicles is primarily hindered by safety concerns. Significant efforts have been devoted to identifying and mitigating unsafe components through testing \cite{zhangPerformanceEvaluationMethod2022,zhaoGeneticAlgorithmBasedSOTIF2023,fahrenkrogEuropeanResearchProject2023}. Scenario-based testing has demonstrated its potential for efficiently evaluating autonomous vehicles under corner case conditions \cite{nalic2020scenario}. Nevertheless, the generation of realistic and plausible corner cases remains a substantial challenge, primarily due to the inherent complexity of physical environments and traffic conditions encountered in the real world. An approach to address the realism challenge involves replaying driving data collected from real-world scenarios; however, this method falls short in creating realistic interaction with av being tested. 

Significant efforts have been made in creating challenging corner cases, broadly classified into two main approaches: data-driven generation and knowledge-based generation. Data-driven models leverage information from collected datasets \cite{yang2023unisim,li2019aads,tan2021scenegen}. For example, NeuralNDE \cite{yan2023learning} employs a Transformer-based network with safety mapping to generate realistic agent behaviors, achieving distribution-level similarity to real-world distributions. STRVE \cite{Rempe_Philion_Guibas_Fidler_Litany_2022} learns a graph-based conditional VAE as traffic prior, optimizing each agent's behavior to provoke collisions with a rule-based AV planner. RealGen \cite{ding2023realgen} uses an encoder-decoder architecture and retrieval-based in-context learning to synthesize realistic traffic scenarios. However, these methods are limited to generating scenarios derived from existing datasets, lacking the capability to produce controllable, specialized scenarios tailored to specific testing objectives.

Knowledge-based generation is another approach that integrates external knowledge into the scenario generation process, reflecting a growing trend in the machine learning field. Klischat and Althoff \cite{klischat2019generating} utilized an evolutionary method to minimize the drivable area by manipulating surrounding vehicles. Shiroshita et al. \cite{shiroshita2020behaviorally} emphasized the importance of diversity and high driving skills within scenarios, proposing a distinct policy set selector within their reinforcement learning method to balance these two aspects. Ding et al. \cite{ding2021semantically} explicitly incorporated domain knowledge by representing it as first-order logic in a tree structure to achieve Semantically Adversarial Generation (SAG). However, these knowledge-based methods generally suffer from limited realisim.

\subsection{Scenario Generation with LLMs}

Multimodal LLMs have seen extensive application in autonomous driving systems since their inception \cite{cui2024survey,tian2024drivevlm,xu2024drivegpt4,cui2024drive}. Specifically, LLMs are increasingly employed for generating diverse and realistic scenarios \cite{chang2024llmscenario,li2024chatsumo,li2024chatgpt}, which are crucial for testing and evaluating autonomous vehicles. For instance, ChatScene \cite{zhang2024chatscene} uses LLMs to generate scenarios from a pre-existing Scenic library \cite{fremont2019scenic}, while ChatSim \cite{wei2024editable} produces photo-realistic 3D driving simulations using external digital assets. However, both methods face limitations in scene diversity: ChatScene is constrained by its fixed library, and ChatSim can only modify existing scenes, unable to create new ones from scratch.

Large language models (LLMs) are also being utilized for generating corner cases. CRITICAL \cite{tian2024enhancing} uses LLMs to refine critical cases by updating scenario configurations for autonomous vehicle training. However, its evaluation is limited to freeway scenarios \cite{krajewski2018highd}, restricting the generalizability across different traffic conditions and road layouts. 
LEADE \cite{tian2024llm} incorporates an LLM-enhanced adaptive genetic algorithm to search for safety-critical scenarios. Nonetheless, the assumption that background vehicles and pedestrians strictly adhere to traffic laws may overlook safety-critical situations resulting from anomalous participant behaviors. CTG++ \cite{zhong2023language} leverages an LLM to transform user queries into a loss function, guiding a diffusion model to generate query-compliant trajectories. However, it is constrained to manipulating agent behaviors within a predefined road map with specific initial conditions. LLMScenario \cite{chang2024llmscenario} employs LLMs to generate brief agent trajectories based on minimal scenario descriptions, thereby facilitating scenario engineering. Nevertheless, it remains restricted to highway scenarios and requires further exploration in more complex environments.

Safety-critical scenario-based testing is a crucial and pressing challenge in the field of autonomous driving, necessitating a vast number of diverse and highly controllable scenarios, a need that remains inadequately addressed. Our work aims to leverage LLMs to facilitate the generation of diverse safety-critical scenarios in a controllable and efficient manner through multiple input modalities, including text, images, and videos.

\section{Method}
\label{sec:method}
In this section, we delineate the key method of AutoScenario, the multimodal LLM-driven tool for converting multimodal input to corner case generation. We begin by introducing concise notation for key components. Then we will introduce the overall pipeline for the whole system, followed by an explanation for the key design of the generalized interpreter. Lastly, we introduce how the corner cases are generated by grouping these components together.

\subsection{Notation}
\begin{figure}[htbp]
  \centering
\includegraphics[width=0.48\textwidth]{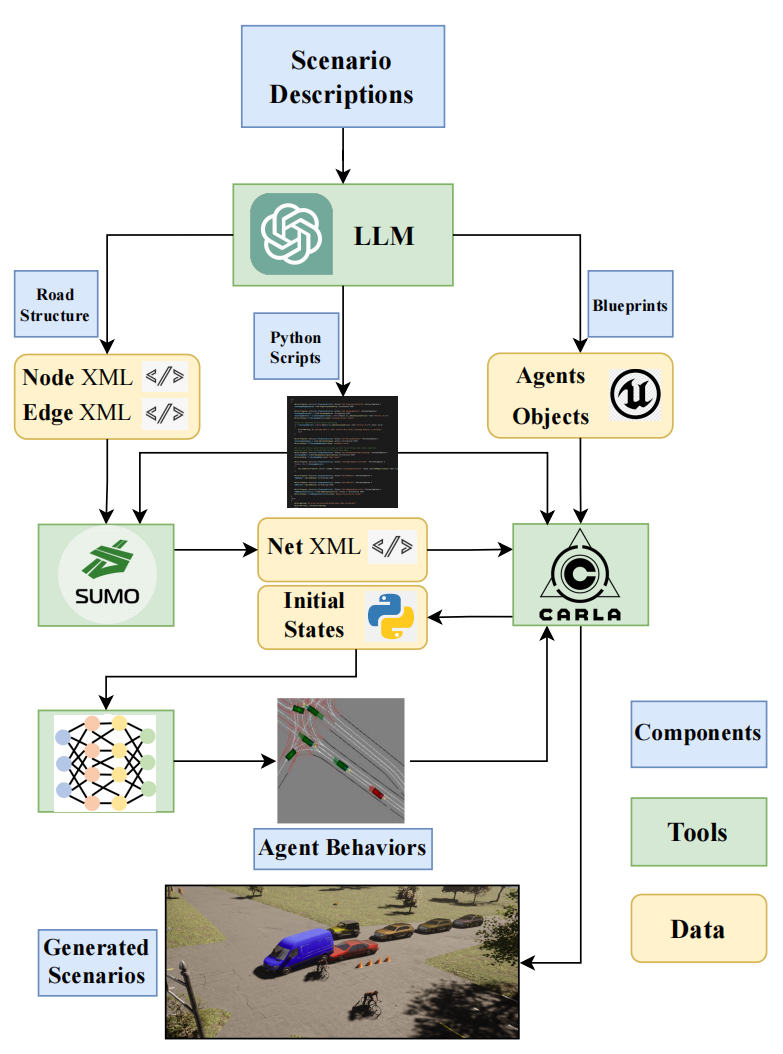}
  \caption{
Tools utilized in AutoScenario: SUMO, CARLA and data-driven models. \label{fig:road}}
\end{figure}

We model the entire process as an encoding-decoding framework, where the input—regardless of modality—is first encoded into a universal, interpretable language space. It is then decoded using multiple codes that direct simulation tools to precisely reconstruct the specific scenes.

Let $S_E$ represent a real-world scenario that encompasses, but is not limited to, elements involved in traffic, such as roads, diverse road users, static objects, traffic signs, and weather conditions. A consistent linguistic description,  $E_l = \{E_{road}, E_{objects}, E_{agents}, E_{weather}\}$ , is constructed with the help of LLMs. This description includes details about the road structure, static objects, agents, and weather conditions. We operate within this consistent space. The mapping from the environment states to the linguistic description can be expressed as:
  $f(S_E) \rightarrow E_l$.

In the decoding process, we leverage several LLM-powered modules to generate the final scenario, with the aid of simulation tools. For instance, the network generator takes in $E_{l}$ with additional domain knowledge $k$ to produce possible lane configurations based on the provided constraints. The agent generator $v$ then generates agent behaviors based on the description $L$, the network structure $n$, and the domain knowledge $k$. Similarly, the object generator $o$ places objects in the environment based on $L$, $n$, and $k$. Finally, the scenario generator $s$ integrates the network, agents, and objects to produce the final scenario $s$, as described by Equation \ref{eq:1}:

\begin{equation}
\label{eq:1}
\begin{aligned}
\underset{k\sim Knowledge}{minimize}& \quad dist(embedding(L), embedding(f(s))\\
	st. \quad L& \sim f(S_E) \\
	n& \sim NetGenerator(L|k) \\
	v& \sim AgentGenerator(L, n|k) \\
	o& \sim ObjectGenerator(L, n|k) \\
	s& \sim ScenarioGenerator(L, n, v, o|k), \\
\end{aligned}
\end{equation}
where $\sim$ denotes sampling from the distribution.

To get the embedding for the objective function, we use the 'text-embedding-ada-002' model \cite{neelakantan2022text} to extract embeddings from the descriptions and calculate the distance between the two embeddings using cosine similarity.

\subsection{Pipeline}
The generation framework is illustrated in Fig. \ref{fig:framework}.
The pipeline accepts multimodal inputs, which includes but not limited to: user request expressed in the text, image taken from random viewpoints, videos from a driving vehicle's perspective.
They are sent through specially designed interpreter powered by LLM to generate standard scenario descriptions, which extracts key components from the input while adding diverse details. See more details in Section \ref{subsec:input}.

Multimodal inputs are preprocessed with tailored attention mechanisms to generate consistent descriptions from the provided real-world information. For short user requests, we expand them to produce more detailed descriptions. For longer texts with a fixed narrative style, such as crash reports, we restructure the information into four perspectives:$E_{road}, E_{static}, E_{agents}, E_{weather}$. For images, we use Chain-of-Thought (CoT) \cite{wei2022chain} to extract layered, risk-related information. For videos, we downsample the footage and utilize memory-based processing to reconstruct the entire network and motion continuity from the input scenario. See detailed prompts in the Appendix.

Road structure plays a crucial role in scenario generation and in identifying risk factors critical to corner case generation. To achieve this, two approaches are employed. In the first approach, $E_l$ generated by interpreters are used to invoke the net generator, which produces the network in XML format. Alternatively, real-world road geometry is retrieved from OpenStreetMap \cite{OpenStreetMapFoundation} using GPS input and converted into the net.XML format. The agent generator creates a set of agents based on $E_{agents}$, which includes a diverse range of road users, such as pedestrians, cyclists, and various types of vehicles including trucks and passenger cars. LLMs are used to place agents within the scenario and assign appropriate speeds. Then, closed-loop simulations are conducted with a data-driven agent model to replicate human-like behaviors and interactions under the given traffic conditions. The object generator, on the other hand, creates static elements such as lane markings, traffic signs, fences, and traffic cones, which remain invariant over time.

\subsection{Tools in the chain}
\label{subsec:tool}
As illustrated in \ref{fig:road}, SUMO, CARLA, and data-driven models are seamlessly integrated with the help of LLM to generate the final scenario.

\textbf{Net Generator} Road geometry is a well structured object that can be represented by graph with nodes and edges. To reduce the error rate for pure generation, we prompt LLM to generate SUMO compatible Node and Edge file defined in XML format, then convert to full graph with domain knowledges, i.e. rules with SUMO tools.  See Fig. \ref{fig:road} for an illustration and more detailed examples in Section \ref{sec:result}. This design is not confined to SUMO or its XML formats. Since road network naturally represents a graph structure, it can be represented by other structured languages \cite{GraphMLFileFormat} and processed by graph tools \cite{hagbergExploringNetworkStructure2008}, compatible to other simulators such as MATSim \cite{MultiAgentTransportSimulation}.

\textbf{ScenarioGenerator} We use LLM to directly generate Python code that controls 3D scenario agents via the CARLA Python API. Blueprints from CARLA’s library are automatically utilized to construct diverse road users and simulate varying weather conditions. Digital Twin Tool is also employed to render the scenario realistically, with all critical risky factors generated automatically through the API.

\textbf{AgentBehavior} We use a trained data-driven behavior model to replicate human-like actions in the given scenario, once the state at a critical moment is generated based on the universal scene description $E_l$.

\section{Experiments}\label{sec:result}

In this section, we evaluate the performance of our AutoScenario framework both quantitatively and qualitatively. First, we examine its ability to generate realistic, diverse, and controllable scenarios. Next, we demonstrate its application in creating safety-critical scenarios from multimodal inputs for AV testing, and additionally, we quantitatively assess the challenges posed by these scenarios through the performance of LLM-based AVs. Finally, we highlight the key components of the framework through an ablation study.

\subsection{Realism, Diversity and Controllability of Generation}\label{subsec:compliance}
\begin{figure*}[htbp] 
  \centering  
\includegraphics[width=0.9\textwidth]{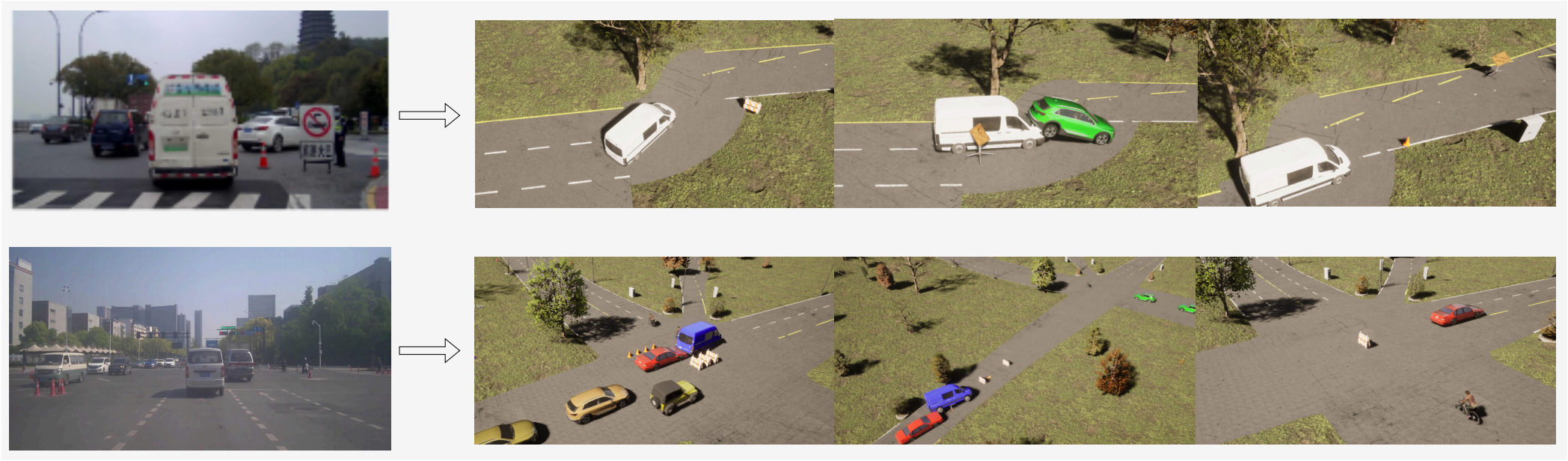}
  \caption{
  Based on the same input from image interpreter, AutoScenario can generate diverse scenarios. \label{fig:image_diverse}}  
\end{figure*}

In our experiments, we observed that with proper prompting, the whole system displays high level of controllability and diversity in generated scenarios. Fig. \ref{fig:image_diverse} listed two samples with image interpreter. The interpreter effectively identified critical scene elements, including road network configurations, vehicle count and color, and obstacles such as construction cones. This information is then seamlessly translated into the generated 3D scene by the components generator and the scenario generator.

To further systematically evaluate the diversity and fidelity of generation, we define a set of metrics to capture the quality of the two main components inside the generation process, both the interpreter and the generator: 
1) \textit{Conformity metrics}, which measure the alignment between the text description and the generated scenario, including network structure, objects, agents, and intermediate codes. Success rates assess the likelihood that these generated codes are correctly recognized by SUMO and CARLA (2) \textit{Diversity metrics}, which evaluate the diversity of scenarios attributes across generated scenarios. Additional experimental details and metric descriptions are provided in the appendix. 

As shown in Table \ref{tab:conformance}, AutoScenario exhibits high success rates in producing meaningful scenarios. Additionally, the system achieves high accuracy in generating the specified number and color of vehicles, as well as the type of road obstacles, according to descriptions generated by the interpreter in most cases. The main causes of failures include incorrect formatting of keywords (e.g., extra ``\#'' characters) in network generation and blueprint name reusing errors (e.g., ``vehicle.omni.vehicle.omni.bus''). Given the complexity of the prompt, these success rates and accuracy levels represent quite favorable results.

Another feature of our system is the diversity and complexity of generated scenarios. To evaluate road network complexity, we calculate the mean and standard deviation of the total number of lanes, edges, and route lengths in the generated road networks. As shown in Table \ref{tab:diversity}, these metrics span a wide range across each scenario set, demonstrating the consistent diversity of the generated outputs. 
\begin{figure}[!h] 
  \centering  
\includegraphics[width=0.48\textwidth]{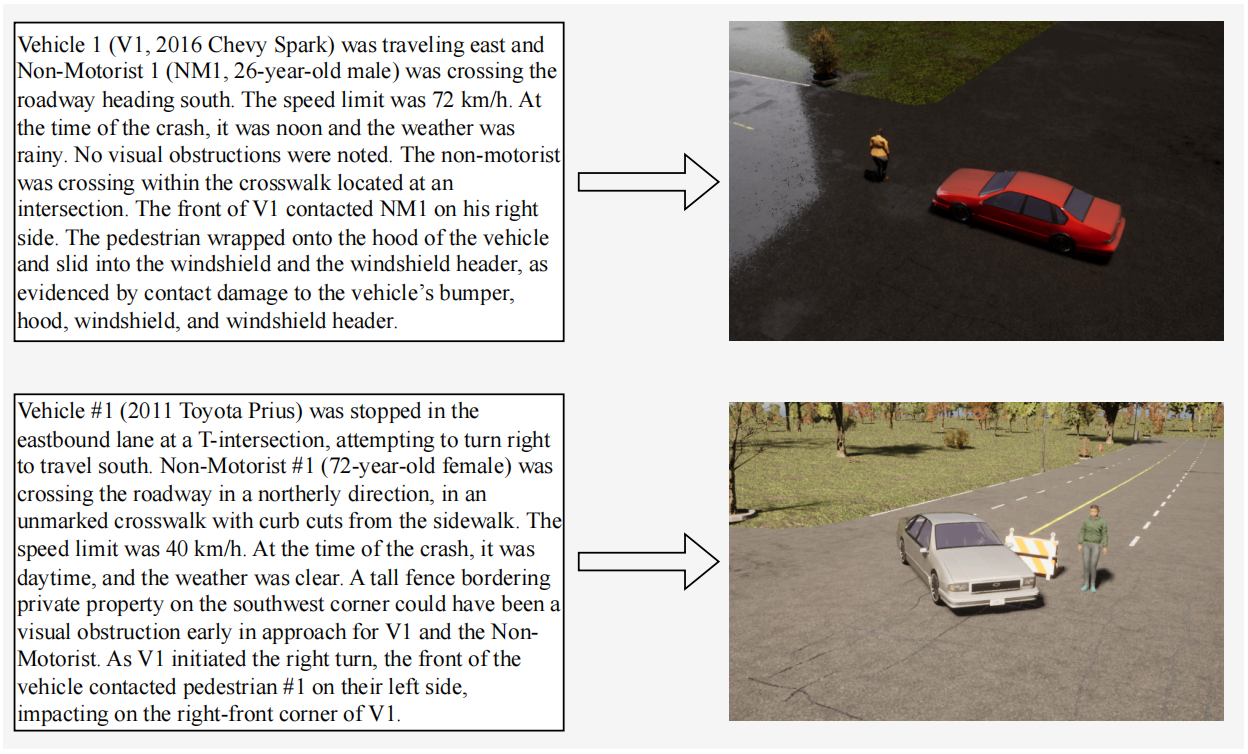}
  \caption{Left: AutoScenario generation using crash reports from NHTSA \cite{NHTSA2023} as input. Right: The scene generated at the moment before accidents.\label{fig:crash_report}}  
\end{figure}

\begin{table}[!h]
\caption{Conformity of command\label{tab:conformance}}
\centering
\footnotesize
\begin{tabular}{|c|c|c|c|}
\hline
Accuracy & General  & Intersection & Construction Zone \\
\hline
Scene Type& 1 & 1 & 1 \\
\hline
Vehicle attributes & 0.9 & 0.65 & 0.67 \\
\hline
Static Objects attributes & 0.93 & 0.9 & 0.96 \\
\hline
Success rate & 0.87 & 0.7 & 0.6 \\
\hline
\end{tabular}
\end{table}

\begin{table}[!h]
\caption{Diversity \label{tab:diversity}of generated scenarios}
\centering
\scriptsize
\begin{tabular}{|c|c|c|c|}
\hline
Scenario & General & Intersection &  Construction Zone\\
\hline
\#Lanes & $6.00 \pm 3.27$  & $11.0.\pm 3.00$ & $11.00 \pm 2.94$ \\
\hline
\#Edges & $5.67 \pm 3.68$  & $7.67\pm 2.43$ & $6.00 \pm 1.63$ \\
\hline
Route Length& $260.70 \pm 114.58$  & $540.04\pm 200.15$ & $420.56 \pm 64.62$ \\
\hline
\#Distance & $58.21 \pm 42.15$ & $6.35\pm 3.36$ & $27.16 \pm 18.01$ \\

\hline
\end{tabular}
\end{table}

\begin{figure}[!h]
\centering 
\includegraphics[width=0.45\textwidth]{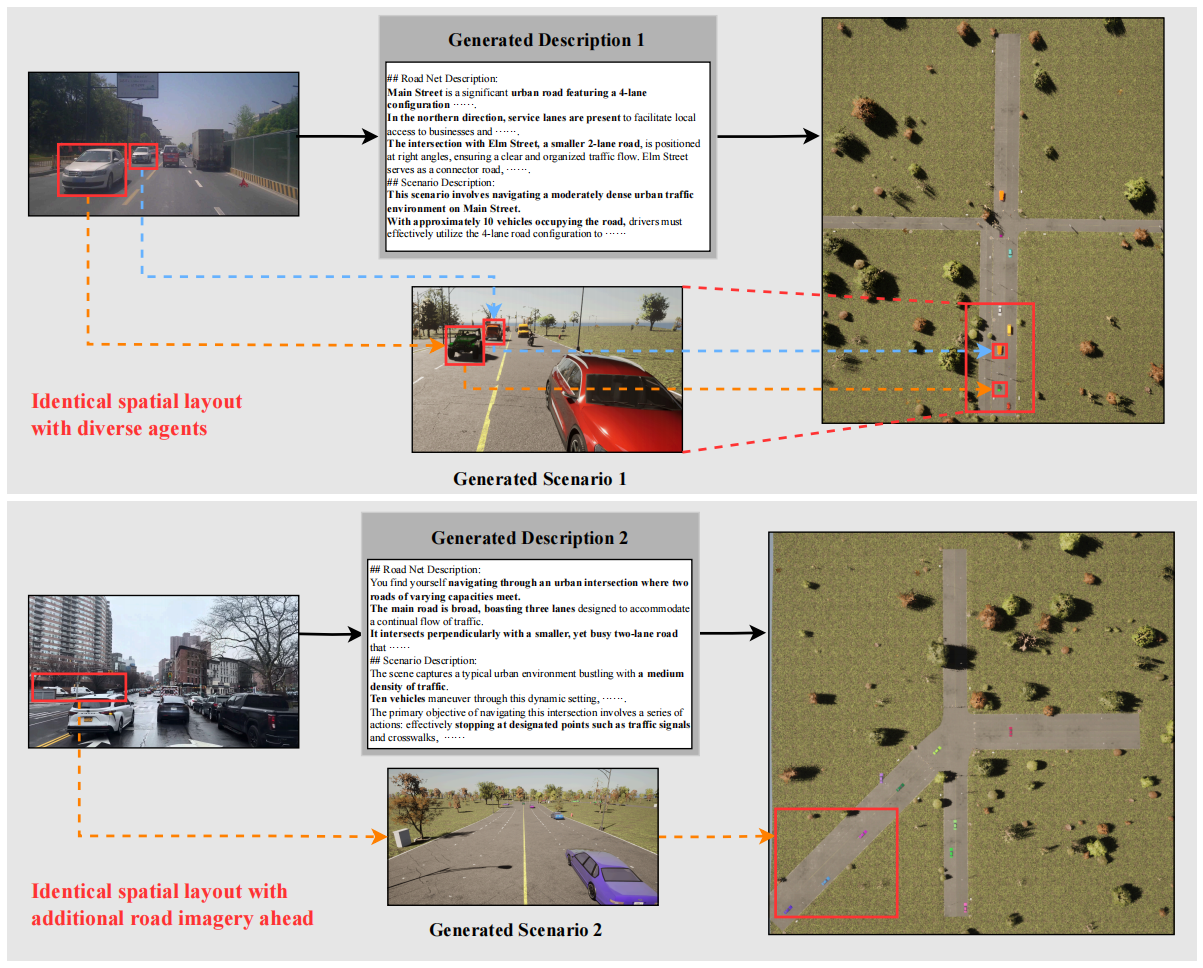}
\caption{Left Column: Real-world scenarios (top-left image from the CODA dataset), representing the first-person perspective of the autonomous vehicle (AV).  Middle Column (Text): Scenario description generated by the image interpreter.  Middle Column (Image): Simulation scenario generated by CARLA and LLM, shown from a perspective different from the overhead view.  Right Column: Overhead-view simulation scenarios generated by CARLA and LLM.\label{fig:img-input}} 
\end{figure}

\subsection{Generation of corner cases using diverse input sources}\label{subsec:input}
Leveraging multimodal interpreters, AutoScenario generates safety-critical scenarios from diverse input sources, including text, images, and videos. The output from AutoScenario is also diversified by highlighting the key components of a conflict scenario while generalizing the rest.

\begin{figure*}[!h]
\centering 
\includegraphics[width=0.95\textwidth]{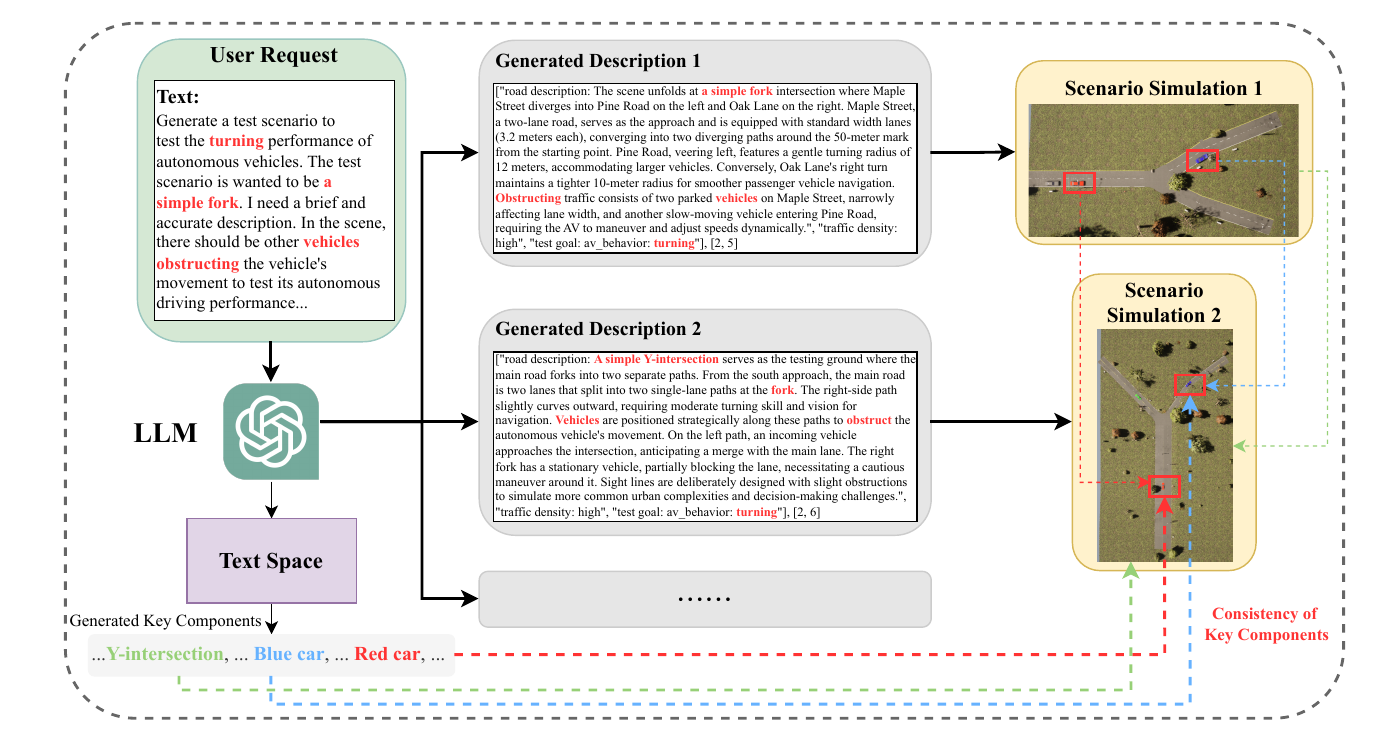}
\caption{AutoScenario Generation Pipeline with User Request: The user specifies the testing requirements (left), and multiple scenario descriptions are generated by the text interpreter (middle). Irrelevant parts of the description are omitted, while key elements are highlighted in bold. On the right, a sampled scenario generated by CARLA and LLM for each description is displayed.
\label{fig:user_request}} 
\end{figure*}

One application of the text interpreter is to reconstruct critical moments leading up to a crash based on crash reports. Two examples of this are shown in Fig \ref{fig:crash_report}, highlighting safety-critical interactions between vehicles and between vehicles and vulnerable road users. Additionally, AutoScenario supports user testing requests that describe scenarios at an abstract level, as demonstrated in Fig in \ref{fig:user_request}.

A properly prompted VLM is used as the image interpreter to convert the input image into a scenario description. It carefully considers the four key perspectives of the scenario. Next, tools in the chain \ref{subsec:tool} are employed to convert the scenario description into a simulation scenario.  During our experiment, we found that the performance of the interpreter was limited in complex scenes—specifically, those with numerous buildings and vehicles—resulting in significant deviations in the extraction of road network features. To address this, we introduced an enhanced prompt that enabled the model to analyze the road network more effectively by leveraging the surrounding buildings and parked vehicles to infer the geometric structure of the road network. This modification led to an improvement in the model's performance in complex scenes. The process of scenario generation is depicted in Fig \ref{fig:img-input}. More details are presented in the appendix.

\begin{figure}[!h]
\centering 
\includegraphics[width=0.49\textwidth]{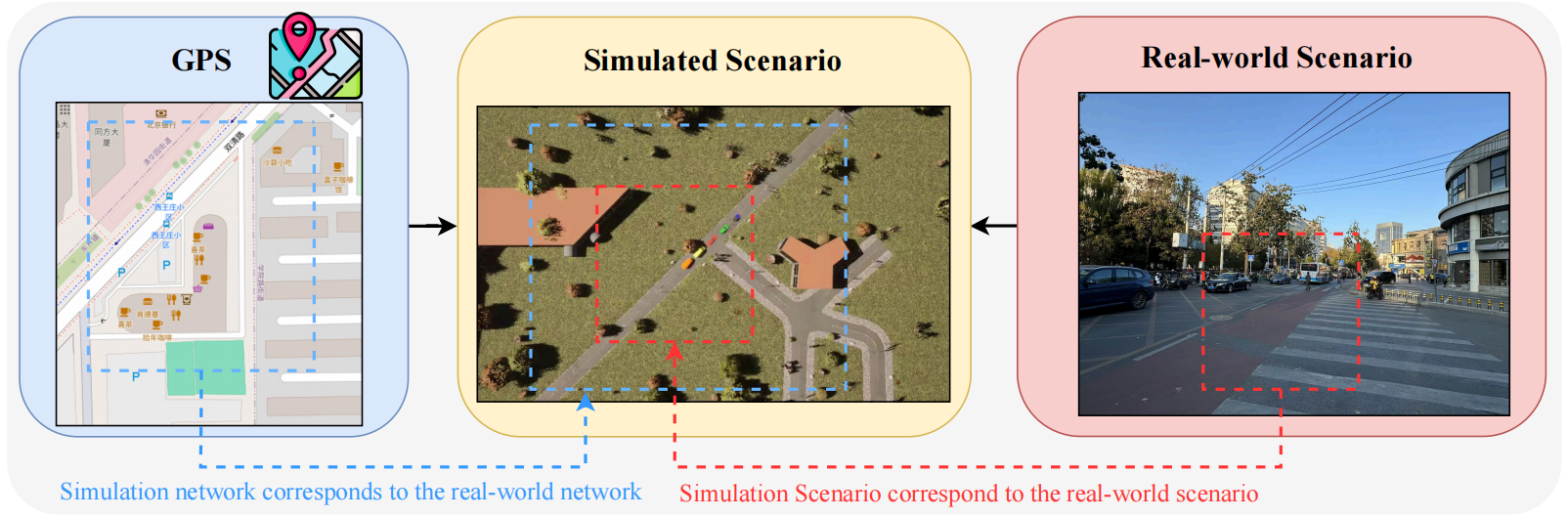}
\caption{AutoScenario generation with additional GPS input.}
\label{fig:gps-input}
\end{figure}

Additionally, when GPS data is provided, we can generate testing scenarios based on the text description while incorporating real-world road structures. This approach enables effective testing for deployment in this region, reflecting realistic traffic conditions, as demonstrated in Fig \ref{fig:gps-input}.

A VLM-based video interpreter is developed to extract road information and environmental features from the input video,  from which a standard scenario description is generated. In the experiment, we observed that the model's estimation of the vehicle's distance while analyzing the driving trajectory in the video was not very accurate. To address this, we introduced a code prompt that calculates the forward distance traveled by the vehicle between two timestamps, using the corresponding images and their depth maps. Compared to the original input, which consisted solely of frames generated from processed videos, our approach also incorporates depth map frames corresponding to these images. With these enhancements, the model's accuracy in estimating the distance traveled in the video, as well as the length and proportions of the road network, has significantly improved. The process of scenario generation is depicted in Fig \ref{fig:video-input}. More details are presented in the appendix.

\begin{figure*}[!h]
\centering 
\includegraphics[width=0.95\textwidth]{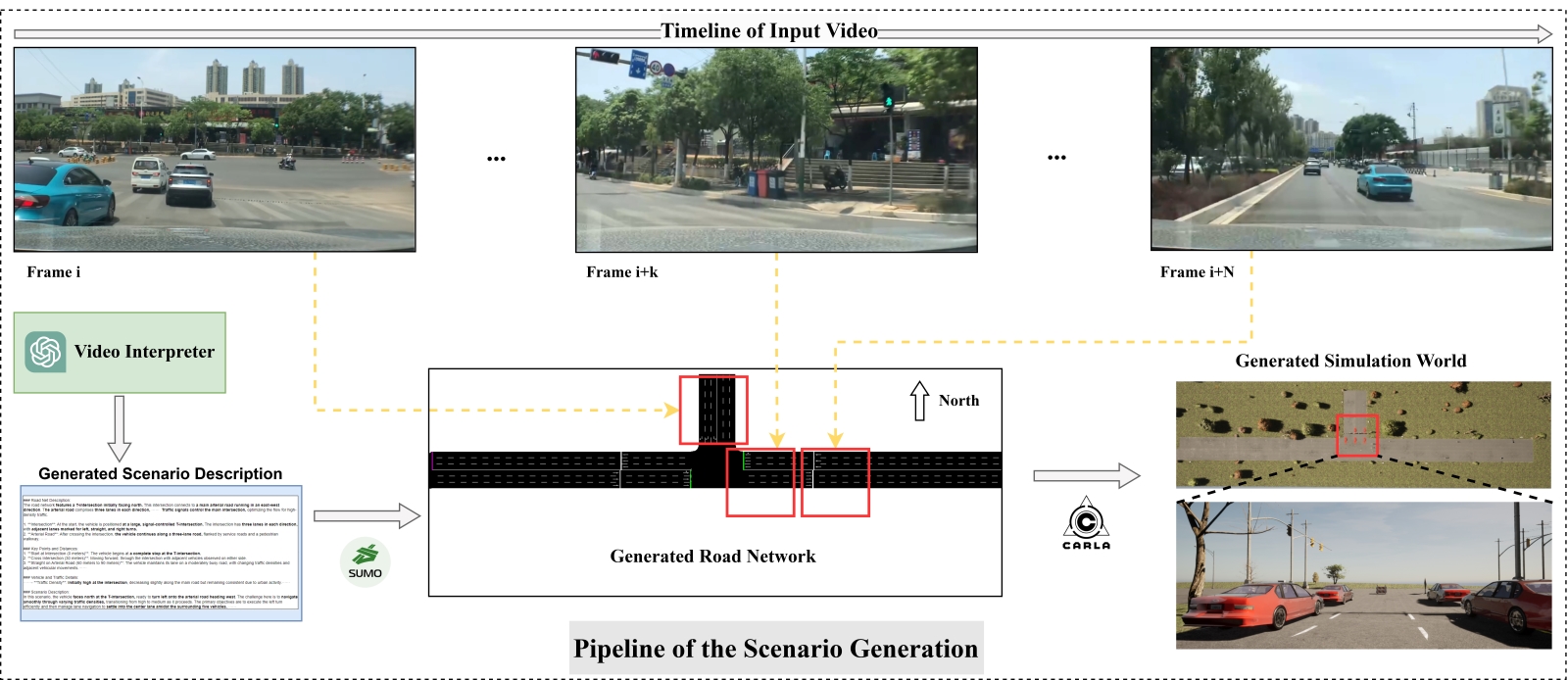}
\caption{Conversion from dashcam video to network file and simulation scenario. Top: Frames extracted from the video. Bottom: Scenario generation process—left: scenario description generated by the interpreter; middle: corresponding network file generated by net generator; right: generated scenario."
\label{fig:video-input}} 
\end{figure*}

Additionally, we perform a quantitative comparison between scenarios generated by AutoScenario and those where traffic vehicles are randomly placed within the same scenario. Both types of scenarios share a network generated from real-world images, which include corner cases. This comparison allows us to assess the effectiveness and realism of AutoScenario-generated scenarios in replicating real-world conditions. For more details on the experimental setup and results, please refer to the appendix.

We have demonstrated scenario generation using various types of inputs, including text, image, and video, along with the key design of their corresponding interpreters. In each case, road information and environmental features are extracted from the respective input. This capability remains a critical component of the AutoScenario system, as it enables the translation of real-world information into a unified text-based format. For additional experiments on the similarity between real-world and generated scenarios, please refer to the appendix.

\subsection{Ablation Study}
\label{sec:ablation}

\begin{table}[!h]
\caption{Ablation study: Change of success rate}
\centering
\begin{tabular}{|c|c|}
\hline
Metrics & Success rate*  \\
\hline
Ours & 0.8 \\
\hline
without interpreter  &0  \\
\hline
without prior knowledge & 0.2 \\
\hline
without reasoning section &0.4 \\
\hline
\end{tabular}
\label{tab:ablation}
\end{table}

The ablation study of AutoScenario is presented in Table \ref{tab:ablation}. In this study, we examine the key design choices across the three main steps: the Interpreter, the Components Generator, and the Scenario Generator, all in relation to the generation process.

After removing the interpreter, which generates detailed narrative descriptions for each scenario, AutoScenario is unable to produce diverse network structures in a single pass, let alone complete the subsequent steps. This highlights the necessity of the multi-stage generation process and underscores the importance of the comprehensive global scenario description.

For the Component Generator, we selected the network generator as the experimental subject. Removing the CoT (Chain-of-Thought) mechanism, a crucial reasoning technique, from its prompt resulted in a noticeable degradation in accuracy. This decline can be primarily attributed to three common issues. First, critical attributes are missing in element definitions, such as the absence of the 'shape' attribute in the 'lane' element, preventing SUMO from correctly interpreting lane configurations. Second, attribute values fall outside the allowed enumeration, as with the 'spreadType' attribute in the 'edge' element, where 'left' was used instead of the valid values ('right,' 'center,' or 'roadCenter'). Third, undeclared attributes are used, like the 'function' attribute in the 'edge' element, which references an incorrect XML Schema. 

In the Scenario Generator, we analyze the success rate when removing the code examples and prior knowledge constraints from the prompt. As shown in Table \ref{tab:ablation}, the success rate of generation drops from 0.8 to 0.2 for the generation process. This decline is expected, as CARLA follows a specific protocol, and without the code examples, the world knowledge embedded in the LLM is insufficient to fully generate the required functionality.

\section{Conclusion and Future work}\label{sec:conclusion}
 
We present a scenario generation framework that integrates LLMs, VLMs, and data-driven models. This is the first system to seamlessly translate multimodal real-world data into simulated scenarios, offering a highly controllable and flexible simulation tool. Transferring and generalizing risky scenarios from the real world to simulators like CARLA, in terms of both objects and behaviors, is a foundational step. In the future, we aim to enhance photorealism using 3DGS or diffusion models.

\clearpage
{
    \small
    \bibliographystyle{ieeenat_fullname}
    \bibliography{main}
}

%
\clearpage
\setcounter{page}{1}
\setcounter{section}{0}
\renewcommand {\thesection}{\Alph{section}}

\maketitlesupplementary
\FloatBarrier
\begin{figure}[!b]
    \centering
    \includegraphics[width=0.95\textwidth]{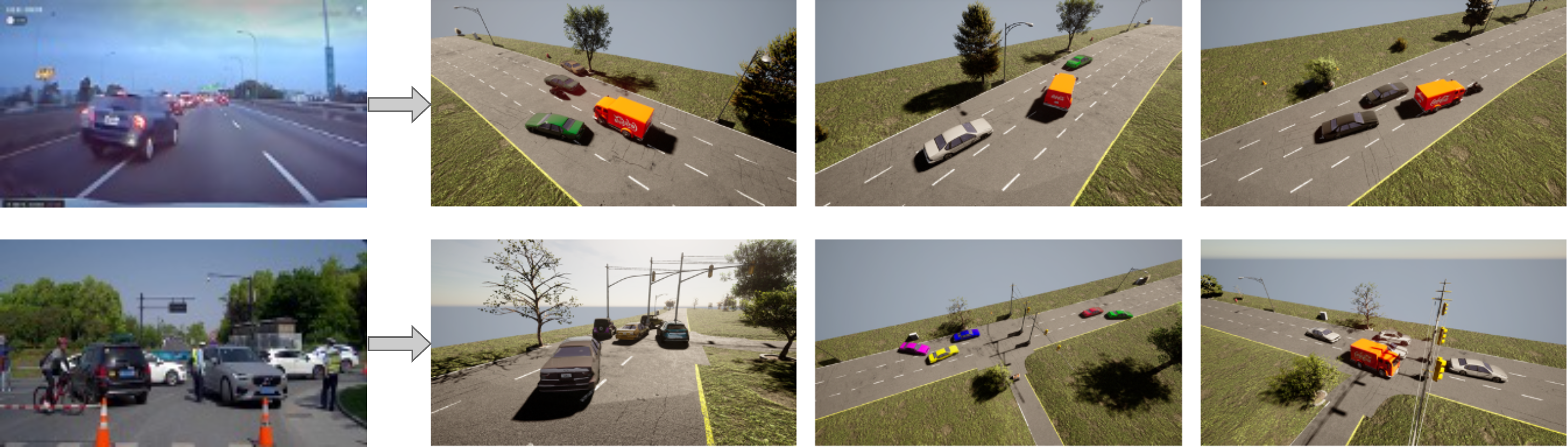}
    \caption{Diverse scenarios generated from the same input image.\label{fig:diverse_img_input}}
\end{figure}
\section{Supplementary Experiments}
\label{sec:appendix_section}

\subsection{Diversity in generated scenarios}
We select three representative scenario types for demonstration: General Scenarios, Intersections, and Construction Zones, as these are areas where corner cases are more likely to occur. For each scenario type, five input images are processed through the AutoScenario pipelines. Each input image is then diversified into 10 distinct testing scenarios. 

To evaluate the diversity of the generated scenarios, we assessed network diversity using metrics such as the number of road users (including vehicles, pedestrians, and cyclist, etc.), the number of static objects (like construction cones or warning signs), the shortest distance between agents in each scenario, and the yaw angles of vehicles generated by AutoScenario, ranging from $-180^\circ$ to $180^\circ$. We calculated the mean and standard deviation of each metric to provide a comprehensive measure of diversity within the scenarios.

As shown in Table~\ref{subtab:diversity}, the standard deviation values indicate a wide distribution range for each metric, reflecting substantial diversity. Compared to the General and Construction Zone scenarios, the Intersection area contains the highest number of agents, which is reasonable since intersections typically have vehicles approaching from multiple directions. The Construction Zone scenarios have the most static objects, consistent with their nature. To simulate corner cases, the shortest distances between agents in all three types of scenarios tend to be around 4 to 5 meters. In the General scenarios, most selected scenarios involve straight roads without intersections, resulting in a mean vehicle yaw angle close to $0^\circ$. In contrast, the other two scenarios involve more complex situations such as lane changes, turns, and merges, leading to an average vehicle yaw angle of approximately $15^\circ$. Additionally, the standard deviation of vehicle yaw angles tends to be around $90^\circ$, which is expected because most road intersections in reality are at $90^\circ$ angles.

\begin{table}[h]
\caption{Diversity \label{subtab:diversity}of generated scenarios}
\centering
\scriptsize
\begin{tabular}{|c|c|c|c|}
\hline
Scenario & General & Intersection &  Construction Zone\\
\hline
\#Agents & $4.19 \pm 2.17$  & $5.63\pm 2.32$ & $3.31 \pm 0.85$ \\
\hline
\#Objects& $2.61 \pm 2.33$  & $2.41\pm 1.13$ & $3.60 \pm 2.98$ \\
\hline
Shortest & $4.40 \pm 2.31$ & $4.29\pm 2.83$ & $5.05 \pm 2.80$\\
\hline
Vehicle yaw & $0.29 \pm 69.77$ & $14.81\pm 93.41$ & $16.46 \pm 90.41$ \\
\hline
\end{tabular}
\end{table}

\subsection{Challenging scenarios created by AutoScenario}
\label{subsec:challenging}

\begin{figure}[!hb]
\centering
\includegraphics[width=0.9\linewidth]{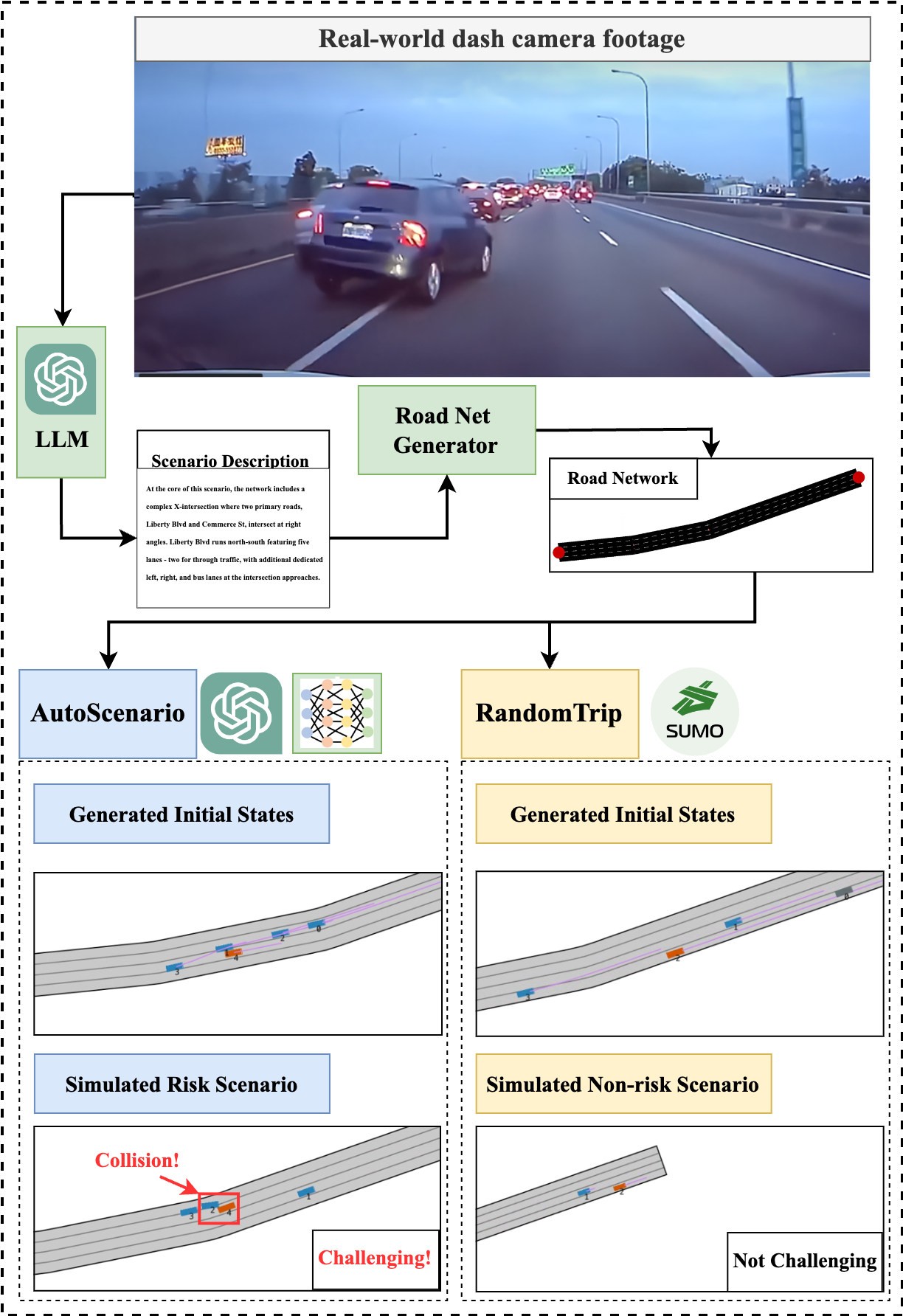}
\caption{Comparison of Scenario Generation Pipelines. Left: A challenging scenario generated by AutoScenario, where the AV and BV are strategically placed based on guidance from input images, and their behaviors are driven by simulation agent models. Right:  Scenarios generated by SUMO, where the AV and BV are spawned randomly in the beginning and BVs are controlled by IDM-based models.\label{fig:challenging-demo}} 
\end{figure}

We compare scenarios generated by AutoScenario with those created by randomly placing traffic vehicles within the same scenario using RandomTrip from SUMO tools. Both types of scenarios are executed on road networks derived from real-world images. This comparison enables us to evaluate the effectiveness and realism of AutoScenario in replicating real-world conditions more accurately. See Fig ~\ref{fig:challenging-demo} for the comparison pipeline.

\begin{table}[h]
\caption{Challenging scenarios generated by AutoScenario for LLM-based AV \label{tab:llm-av}}\centering
\begin{tabular}{|c|c|c|}
\hline
Scenario & Ours & RandomTrip  \\
\hline
Route completion $\downarrow$ & $0.86 \pm 0.26$  & $0.92 \pm 0.24$  \\
\hline
Driving score  $\downarrow$ &$65.24 \pm 16.43$ & $72.87 \pm 19.24$ \\
\hline
Total score  $\downarrow$ &$59.47 \pm 25.25$  & $69.66 \pm 26.15$  \\
\hline
Use Time(s) & $84.09 \pm 44.62$  & $107.96 \pm 49.86$\\
\hline
Success rate $\downarrow$ & $0.76 \pm 0.44$  &  $0.88 \pm 0.33$  \\
\hline
Collision rate $\uparrow$ & 0.2 & 0.08 \\
\hline
\end{tabular}
\end{table}

\newpage
The performance of an AV serves as an indicator of the difficulty level of the generated scenarios. Lower performance suggests that the scenarios are more challenging. To evaluate the AV's effectiveness, we utilize widely adopted performance metrics that account for driving sophistication and task completion levels. The driving score is calculated as a weighted combination of ride comfort, driving efficiency, and driving safety. The route completion value is defined as the ratio of the distance traveled by the driver agent to the total length of the predefined route. The total score is calculated by multiplying the driving score by the route completion.  For more details, see Limism++. In our experiments, we utilized five different road networks, with each network generating five distinct initial vehicle positions using both our proposed method and RandomTrip. Beyond the evaluation metrics mentioned earlier, we also recorded the number of collisions across these 25 experiments to determine the collision rate metric. As shown in Table \ref{tab:llm-av}, AutoScenario generates more challenging scenarios compared to RandomTrip.

\begin{figure}[!h]
\centering
\includegraphics[width=0.9\linewidth]{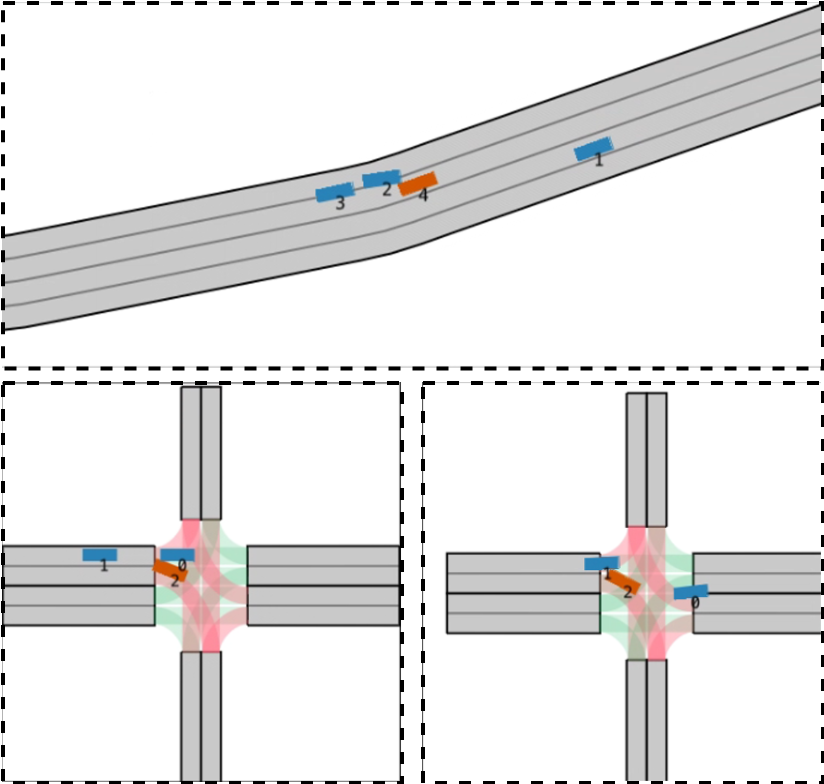}
\caption{AV collides with other vehicles following multiple simulation steps from initial states. The orange vehicle represents the LLM-based autonomous vehicle, while the blue vehicles are simulated agents driven by data-driven models. All vehicles are initialized using the components generators.\label{fig:challenging-case}}
\end{figure}

Three examples of challenging scenarios generated by AutoScenario are illustrated in Fig.~\ref{fig:challenging-case}. As shown, AutoScenario generates challenging scenarios near curved roads (top) and intersections (bottom), driven by a combination of factors: carefully designed initial states and the interactive behaviors of agents. On one hand, it incorporates elements such as BV lane changes and AV turns at intersections, reflecting risky situations observed in real-world driving. On the other hand, by integrating with data-driven simulated agents, AutoScenario creates safety-critical scenarios, enabling interactive and robust testing of AVs.

\subsection{Similarity between generated scenario and original input}

\begin{figure}[!htbp]
    \centering
    \includegraphics[width=0.98\linewidth]{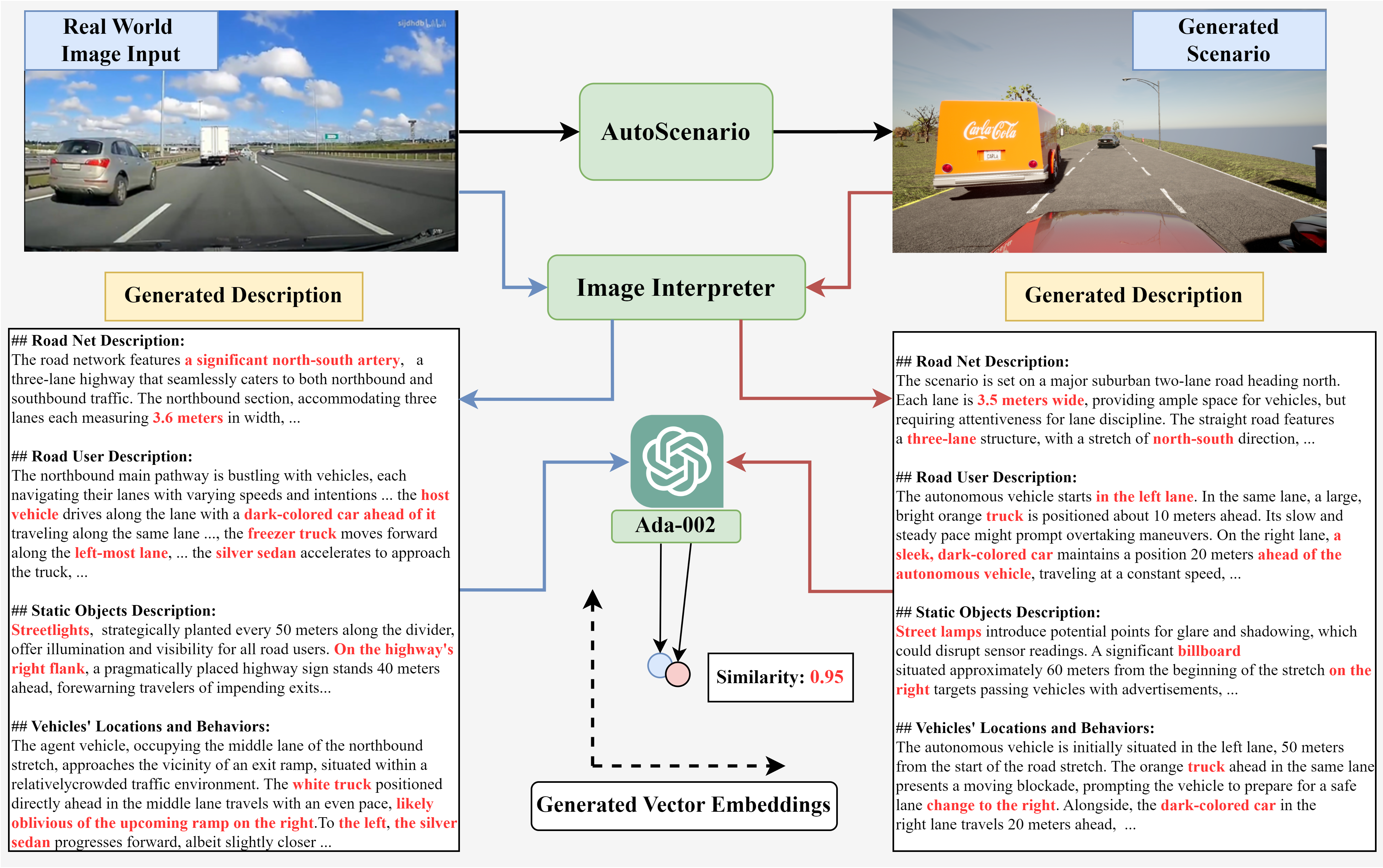}
    \caption{Left: description generated from real-world image. Right:description generated from scenario generated from AutoScenario.
\label{fig:sim-demo}} 
\end{figure}

To validate the similarity between generated scenario and original input image,  a pipeline is designed to automated compare both in the text space generated by the VLM interpreter. See  Fig ~\ref{fig:sim-demo} for a detailed example.  In addition to comparing general scene descriptions, we further break them down into main component descriptions and conduct a detailed comparison for each. Moreover, leveraging the ability to manipulate scene views in CARLA, we selected two distinct perspectives for the generated simulated scene: the bird's-eye view (BEV) and the ego vehicle view. These perspectives were used for cross-checking the similarity between the generated scene and the input image.

\begin{table}[!h]
\caption{Similarity between original input and generated scenarios with ego vehicle view and bird's eye view \label{tab:sim}}\centering
\begin{tabular}{|c|c|c|c|c|}
\hline

Scenario & Ego car view & BEV \\
\hline
Overall scene  & $0.9332 \pm 0.0154$  & $0.9431 \pm 0.0075$  \\
\hline
Net &$0.9065 \pm 0.0157$ & $0.9222 \pm 0.0196$  \\
\hline
Road User &$0.9102 \pm 0.0158$  & $0.9049 \pm 0.0173$  \\
\hline
Static object & $0.9053 \pm 0.0155$  & $0.9004 \pm 0.0075$ \\
\hline
Vehicle behavior& $0.9295 \pm 0.0399$  & $0.8565 \pm 0.0955$ \\
\hline
\end{tabular}
\end{table}
As shown in Table~\ref{tab:sim}, the generated scenarios exhibit a high degree of similarity compared with original inputs in the universal text space while maintaining diversity in specific scenario details. 

In the analysis of separate description sections from the ego car's perspective, the vehicle behavior section shows the highest similarity. This could be attributed to the prominence of vehicles in the image from the car's viewpoint, as they occupy the majority of the visual space. Conversely, in the analysis of separate description sections from the bird's-eye view (BEV), the vehicle behavior section exhibits the lowest similarity. This may be due to the overhead perspective, where the road network and broader scene occupy most of the visual space, and vehicles appear very small, making their behavior more challenging to describe accurately.

\subsection{Online learning for AV with corner case}

For the risk scenarios identified through experiments in section ~\ref{subsec:challenging}, we extract the prompt inputs for collision scenarios and the corresponding decision outputs from the LLM. These are then used as examples to refine the prompts for the LLM-based AV. Following each collision example, we incorporate human suggestions, such as decelerating earlier or switching to a safer lane, to improve decision-making and safety. We then retest the AV in exactly the same 25 scenarios as those in section ~\ref{subsec:challenging}. See table ~\ref{tab:llm-av2}, by emphasizing the corner cases, the performance gets improved in general.

\begin{table}[!h]
\caption{Improving AV performance with corner cases \label{tab:llm-av2}}\centering
\begin{tabular}{|c|c|c|}
\hline
Scenario & AV w/o hints & AV with hints \\
\hline
Route completion $\uparrow$ & $0.86 \pm 0.26$  & $0.96 \pm 0.19$  \\
\hline
Driving score  $\uparrow$ &$65.24 \pm 16.43$ & $70.71 \pm 19.08$ \\
\hline
Total score  $\uparrow$ &$59.47 \pm 25.25$  & $70.06 \pm 21.01$  \\
\hline
Use Time(s) & $84.09 \pm 44.62$  & $141.80 \pm 90.21$\\
\hline
Success rate $\uparrow$ & $0.76 \pm 0.44$  &  $0.88 \pm 0.33$  \\
\hline
Collision rate $\downarrow$ & 0.2 & 0.12 \\
\hline
\end{tabular}
\end{table}

\section{Prompt Examples} 
\label{sec:prompts}
\begin{figure}[!ht]
\includegraphics[width=0.98\linewidth]{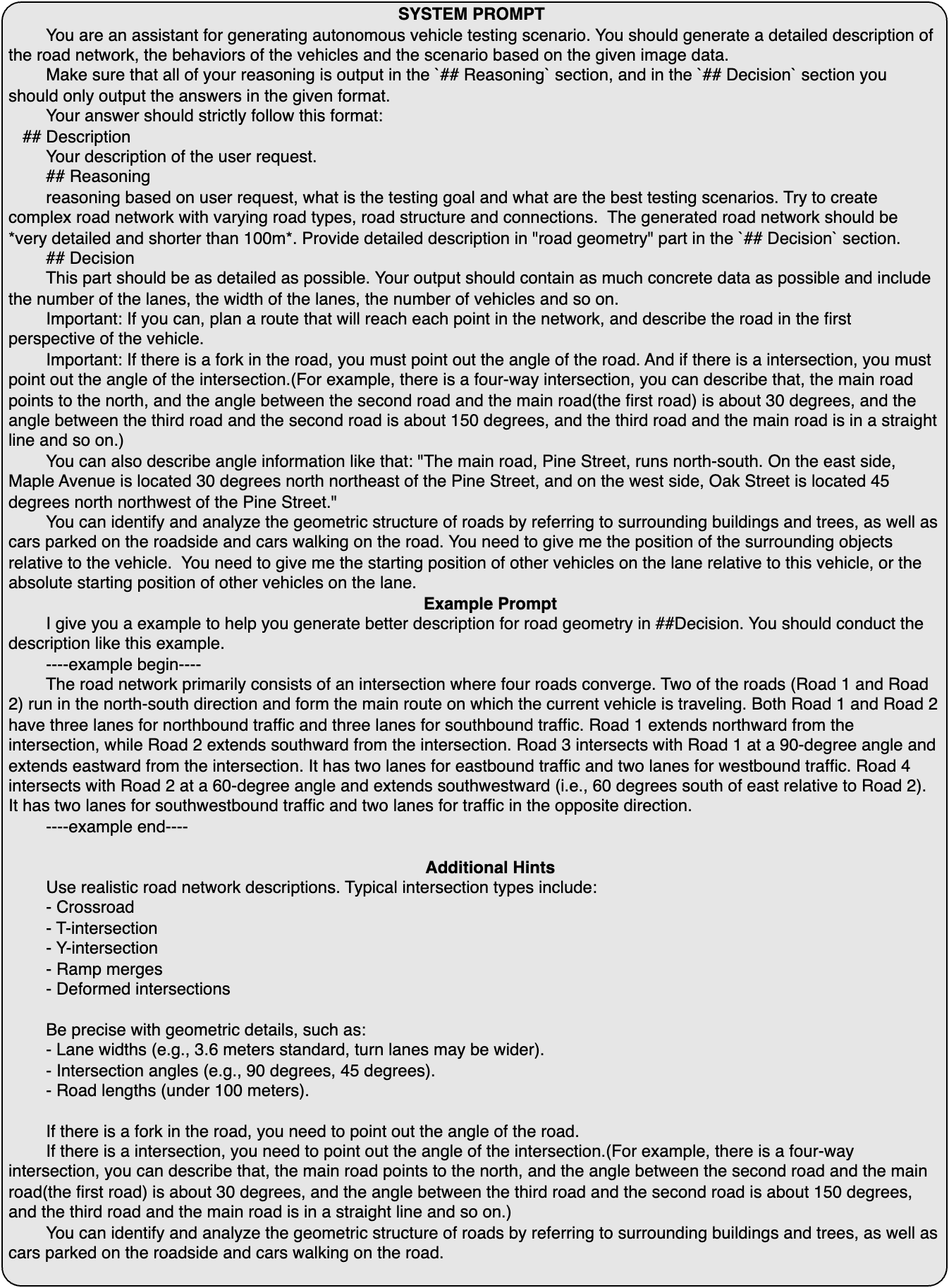}
    \caption{Prompt for VLM Interpreter\label{fig:vlm-prompt}} 
\end{figure}

\begin{figure}[!htbp]
    \centering
\includegraphics[width=0.85\linewidth]{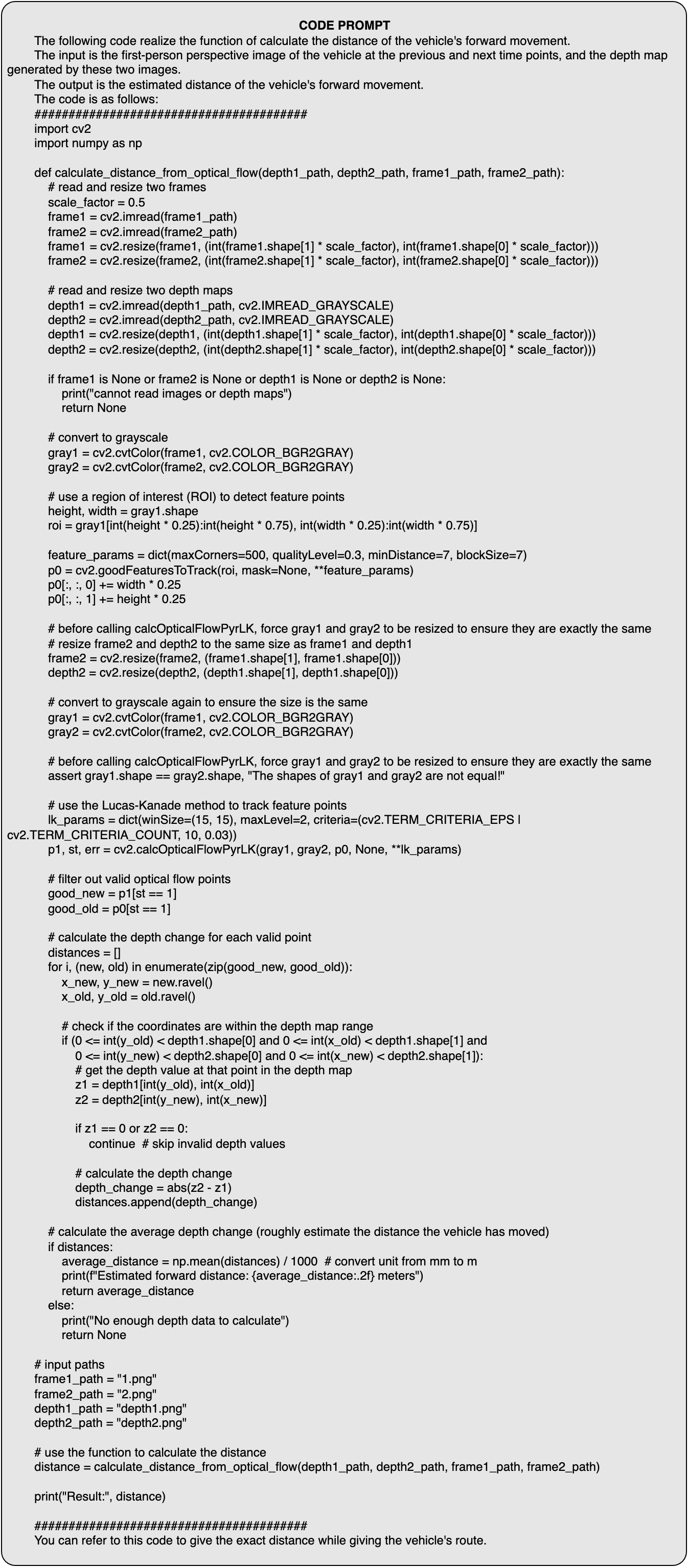}
    \caption{Prompt for Video Inpreterr\label{fig:prompt-video}} 
\end{figure}

\begin{figure}[!htbp]
    \centering
\includegraphics[width=0.85\linewidth]{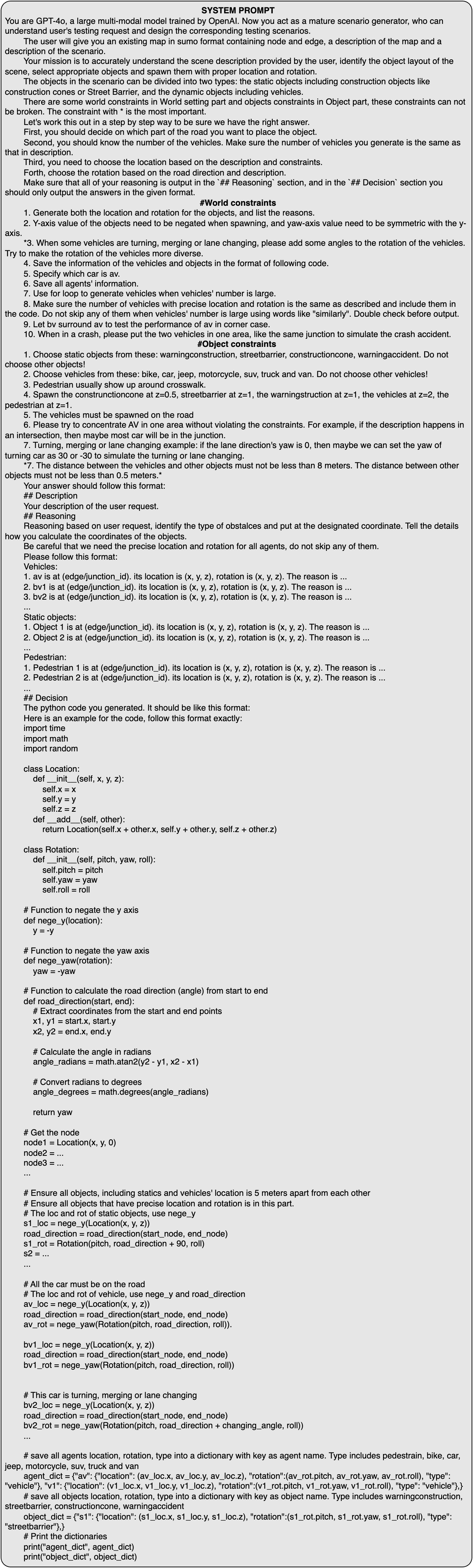}
    \caption{Prompt for agents and objects generator\label{fig:prompt-obj}} 
\end{figure}

\begin{figure}[!htbp]
    \centering
\includegraphics[width=0.85\linewidth]{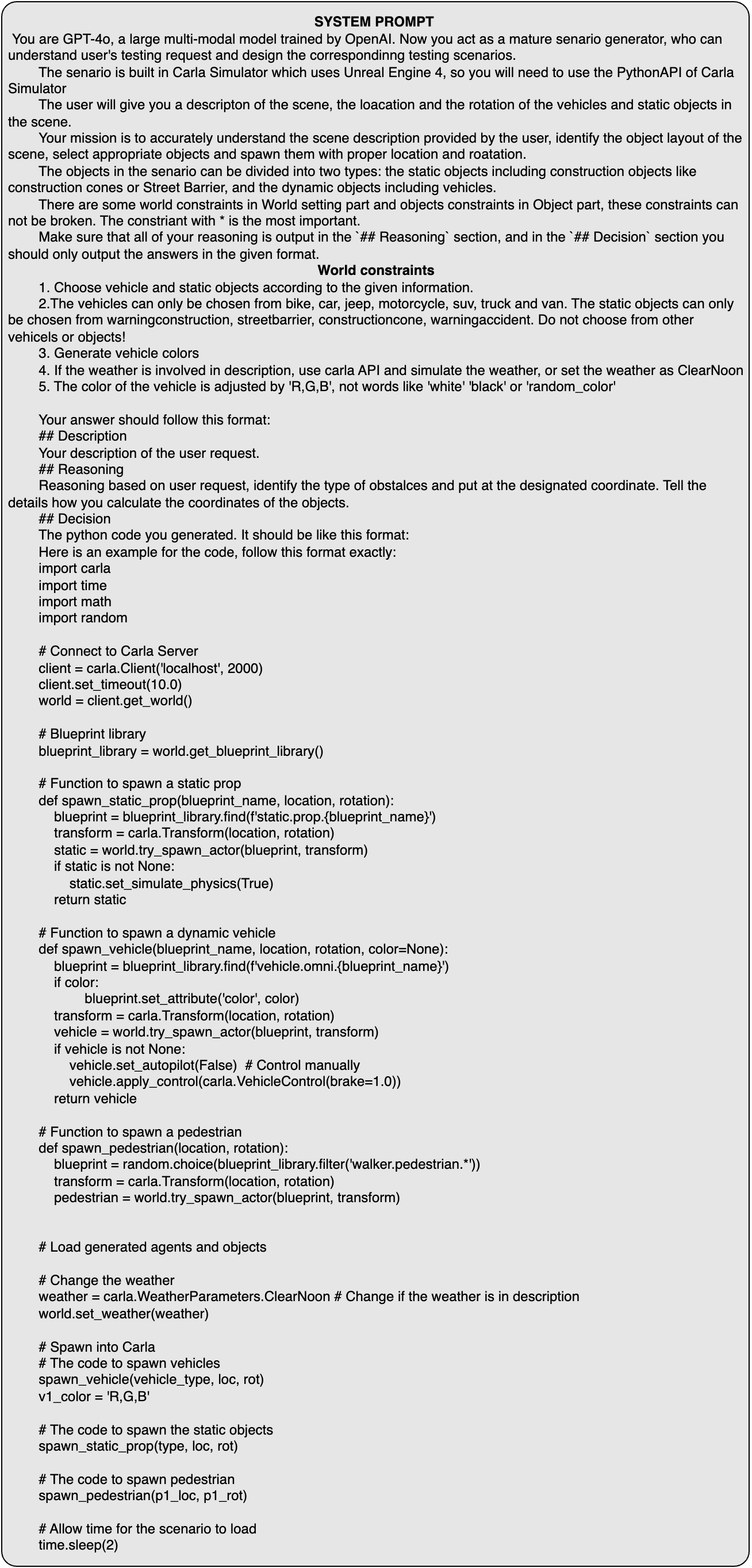}
    \caption{Prompt for scenario generator \label{fig:prompt-scene}} 
\end{figure}

We carefully designed prompts for each component in AutoScenario to fully leverage the capabilities of multimodal LLMs. Overall, each prompt comprises several components: a system prompt summarizing the task, detailed steps to guide the generation process, including constraints and examples (narrative descriptions with code snippets), and the specified format for the desired output. Fig ~\ref{fig:vlm-prompt} and Fig ~\ref{fig:prompt-video} demonstrates the prompt snippets used in the VLM interpreter and video interpreter. For the video interpreter, in addition to prompts similar to those used in the VLM Interpreter, it incorporates code snippets to assist in tracking the ego vehicle's movement. Fig ~\ref{fig:prompt-obj} illustrates the prompt used for agents and objects generator in AutoScenario, while Fig ~\ref{fig:prompt-scene} demonstrate the prompt designed for scenario generator.

Based on our experiments and quantitative evaluations conducted in the ablation study, prompts are shown to significantly enhance the reasoning and generation capabilities of LLMs.\\

\section{Experiment details}
All our experiments are conducted on one NVIDIA RTX 3090, leveraging the online version of GPT-4 as the multimodal LLM alongside a pretrained data-driven model for simulating agent behavior.

\section{More examples}

\begin{figure*}[!h]
  \centering  
\includegraphics[width=0.8\textwidth]{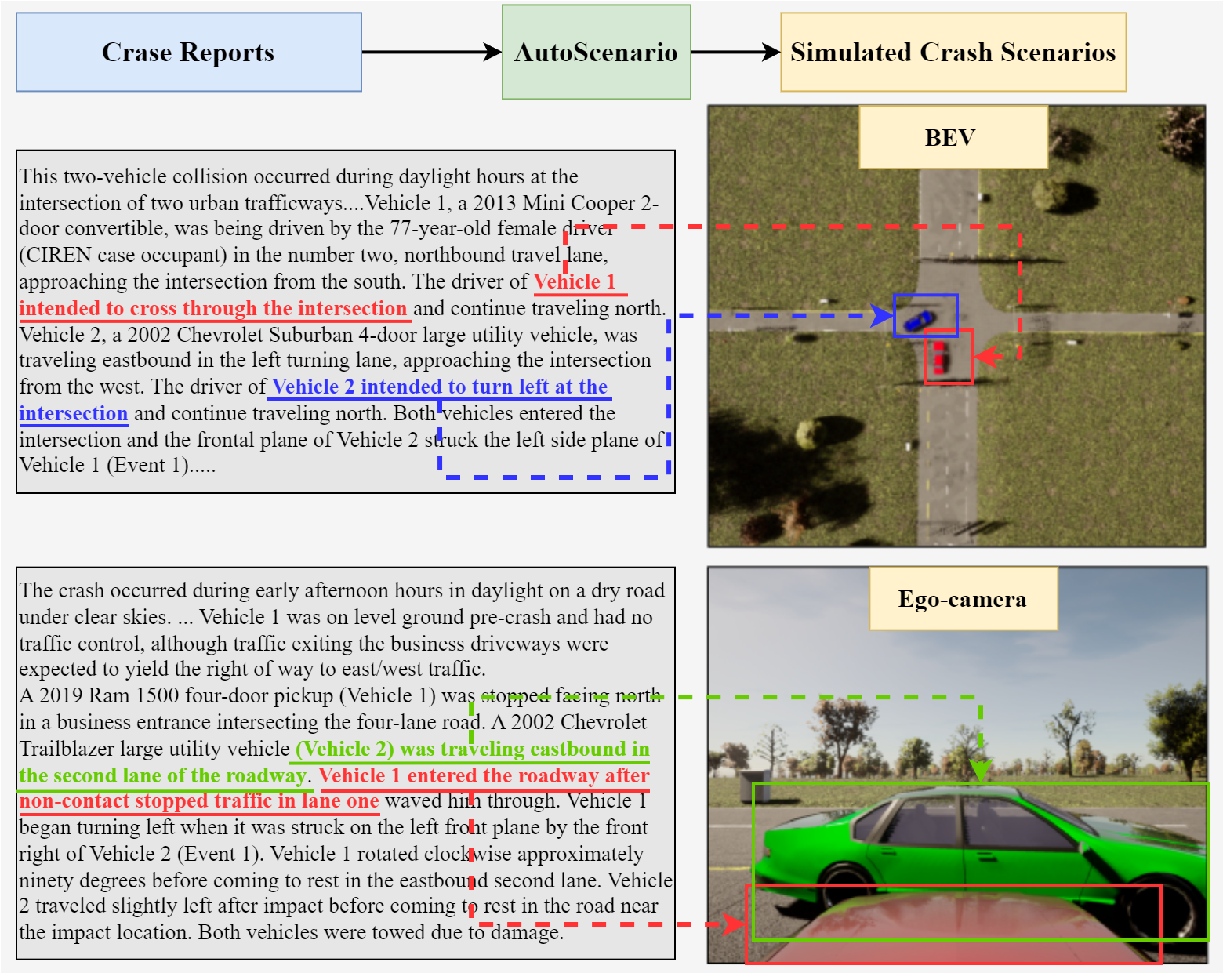}
\caption{AutoScenario using crash report as input\label{fig:prompt_comb}}  
\end{figure*}

Here, we present two additional results derived from crash reports involving conflicts between vehicles, shown in Fig ~\ref{fig:prompt_comb}.

\end{document}